\documentclass[shortpaper]{clv3}

\usepackage{hyperref}
\usepackage{xcolor}
\definecolor{darkblue}{rgb}{0, 0, 0.5}
\hypersetup{colorlinks=true,citecolor=darkblue, linkcolor=darkblue, urlcolor=darkblue}

\usepackage{graphicx}
\usepackage{siunitx}
\usepackage{subcaption}
\usepackage{amsmath} 
\usepackage{fancyvrb}
\usepackage{enumitem}

\DeclareMathOperator*{\argmax}{arg\,max}

\newcommand{\exampletext}[1]{\texttt{\small #1}}
\newcommand{\eos}{{\tt \textless eos\textgreater}}
\newcommand{\self}{{\tt \textless self\textgreater}}
\newcommand{\sil}{{\tt sil}}
\newcommand{\N}{\mathbb{N}}
\newcommand{\M}{\mathbb{M}}
\newcommand{\X}{\mathbf{X}}
\newcommand{\Z}{\mathbf{Z}}
\newcommand{\Y}{\mathbf{Y}}

\usepackage{pifont}
\newcommand{\cmark}{\ding{51}}%
\newcommand{\xmark}{\ding{55}}%

\issue{55}{1}{2019}

\dochead{A Character-Level Approach to the Text Normalization Problem Based on a New \hbox{Causal Encoder}}

\runningtitle{A character-level text normalization model}

\runningauthor{Adri\'an Javaloy \& Gin\'es Garc\'ia}

\jname{}
\jinfo{A character-level text normalization model}
\renewcommand{\footmark}{}
\runningauthor{}

\begin{document}

\author{Adri\'an Javaloy Born\'as \thanks{T\"ubingen, Germany. \href{mailto:adrian.javaloy@tuebingen.mpg.de}{ajavaloy@tuebingen.mpg.de}}}
\affil{Max Planck Institute for \hbox{Intelligent Systems}}

\author{Gin\'es Garc\'ia Mateos \thanks{Murcia,~Spain.~ \href{mailto:ginesgm@um.es}{ginesgm@um.es} }}
\affil{University of Murcia}
\maketitle

\begin{abstract}
    Text normalization is a ubiquitous process that appears as the first step of many Natural Language Processing problems. However, previous Deep Learning approaches have suffered from so-called silly errors, which are undetectable on unsupervised frameworks, making those models unsuitable for deployment. In this work, we make use of an attention-based encoder-decoder architecture that overcomes these undetectable errors by using a fine-grained character-level approach rather than a word-level one. Furthermore, our new general-purpose encoder based on causal convolutions, called Causal Feature Extractor (CFE), is introduced and compared to other common encoders. The experimental results show the feasibility of this encoder, which leverages the attention mechanisms the most and obtains better results in terms of accuracy, number of parameters and convergence time. While our method results in a slightly worse initial accuracy (92.74\%), errors can be automatically detected and, thus, more readily solved, obtaining a more robust model for deployment. Furthermore, there is still plenty of room for future improvements that will push even further these advantages.
\end{abstract}

\section{Introduction}
\label{subsec:text-normalization}
\label{subsec:goals}


The research in natural language processing (NLP) has traditionally focused in the resolution of the \emph{big problems}, such as automatic translation, understanding, summarizing and text generation. However, there are plenty of not so well-known problems that are often overlooked, despite being as hard to grasp as the first ones. In particular, the problem of {text normalization} is one of such cases. It can be defined as: given an arbitrary text, transform it into its {normalized form}. This normalized form depends on the context we are working on. For example, in the context of {text-to-speech} (TTS) systems ---which is the objective of this article--- normalizing a text means writing it as it should be read, e.g.:
\begin{align*}
	\exampletext{I have \$20} &\rightarrow \exampletext{I have twenty dollars} \\
	\exampletext{It happened in 1984} &\rightarrow \exampletext{It happened in nineteen eighty four} \\
	\exampletext{He weights 50kg} &\rightarrow \exampletext{He weights fifty kilograms}
\end{align*}

\begin{figure}[hbtp]
    \centering
    \includegraphics{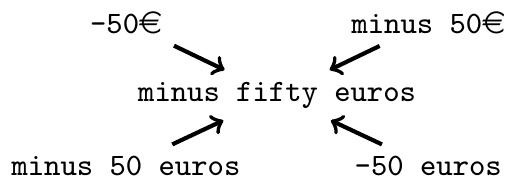}
    \caption{An example of equivalent phrases.}
    \label{fig:cloud}
\end{figure}

At first glance, this problem might seem trivial and rather unimportant, but nothing could be further from the truth. Normalizing text is an ubiquitous process, present in most of the NLP problems. The reason is that normalizing the input as a first step significantly decreases the complexity of those problems, by the fact that equivalent phrases ---yet differently written--- end up being exactly the same phrase, as shown in \hbox{Figure \ref{fig:cloud}}. WaveNet \cite{DBLP:conf/ssw/OordDZSVGKSK16} is an example of these systems, where a generative model for TTS is trained with normalized text as input.

Despite its apparent simplicity, this problem entails a serious challenge. Data-driven approaches, specifically {Deep Learning}, deserve a special mention since: \hbox{(1) there} exists a general belief that Deep Learning can solve any kind of problems, and \hbox{(2) it} is the framework used in this article. Text normalization gathers three features that make it challenging for this kind of techniques, as it has been discussed by \citet{DBLP:journals/corr/SproatJ16}. In short, these features are:
\begin{itemize}
	\item Non-trivial cases (i.e., those whose output and input differ) are sparse.
	\item It is {context-dependent}, for example, a normalized date could change depending on the local variant of the language.
	\item There is no natural reason for building a text normalization database. Everyone knows that \exampletext{2} means \exampletext{two}.
\end{itemize}

A number of different models have been developed to tackle this problem. The first attempts date back to the times when researches were developing the first full TTS systems, as described by \citet{DBLP:journals/corr/SproatJ16}. The systems based on traditional techniques include finite-state automatons as well as finite state transducers \cite{DBLP:journals/nle/Sproat96}. The usage of these models has the advantage of being well-known techniques that work (and fail) as expected; yet, these solutions need to be hand-crafted from scratch for each language, suffering a lack of flexibility (which translates into an increase of production cost).
	
Nowadays, researchers are moving towards Deep Learning models, that try to learn how to solve the problem from the data itself \cite{DBLP:journals/corr/SproatJ16}. However, the amount of information these models require to work well can be prohibitive. In cases where the target language is a low-resourced one, that is, a language for which little data is available, rule-based solutions have been attempted \cite{47344}, as well as Deep Learning models that make use of data augmentation techniques to compensate the lack of data \cite{DBLP:conf/aclnut/IkedaSM16}. In particular, the model described in this article is quite similar to the one proposed here, except for the encoder and other minor tweaks.

The models proposed by \citet{DBLP:journals/corr/SproatJ16} require special attention. They were based on Deep Learning techniques and, in each time-step, they read a character and produce an entire word, so they were character-based at the input level, and word-based at the output. These models obtained a really high accuracy performance (one of them achieving a \SI{99.8}{\percent} on the English test set). Unfortunately, they suffered from {undetectable errors}, that is, errors that cannot be detected by just looking at the output; for example, transforming \exampletext{I'm 12} into \exampletext{I am thirteen}. We suspect that these errors occur as a consequence of using recurrent word-level models.


The present approach has been designed with two goals in mind. The first one is offering a solution for the text normalization problem that exclusively uses neural networks, taking advantage of the benefits of using data-driven solutions. Furthermore, a secondary goal is to introduce convolutional components in this neural model, substituting its recurrent counterparts and, thus, speeding up the whole process. Moreover, proving the usefulness of such convolutional architecture would help to push even further the idea that {Convolutional Neural Networks} (CNN) can be used outside of a computer vision framework.

The main contributions of this work are as follows: \hbox{(1) proposal} of a character-based approach for the text normalization problem which does not suffer from unsolvable and undetectable errors;
\hbox{(2) a} new general-purpose encoder based on causal convolutions, the Causal Feature Extractor (CFE), is introduced and tested; and \hbox{(3) a} variation of the traditional attention mechanisms is introduced, in which a context matrix is generated, instead of a context vector.

\section{Materials and Methods}

\subsection{Dataset}

As stated in \hbox{Section \ref{subsec:text-normalization}}, it can be challenging to obtain a valid database of normalized text. Fortunately, a huge database was built and released to the whole Machine Learning community thanks to \citet{DBLP:journals/corr/SproatJ16}.

This database was shaped for their word-level model and, therefore, it requires some preprocessing before being suitable for a character-level approach. Particularly, each entry on the original database is a pair of words (or a special symbol) plus an additional column describing its semiotic class, as shown in \hbox{Figure \ref{fig:dataset-orig}}. In order to use a character-level approach, each row needs to be composed of all the words pertaining to the same phrase, and information regarding each individual word (such as its semiotic class) has to be discarded.

\begin{figure}[hbtp]
    \centering
    \begin{Verbatim}[frame=single]
"Semiotic Class","Input Token","Output Token"
"PLAIN","Rosemary,"<self>"
"PLAIN","is","<self>"
"PLAIN","a","<self>"
"PLAIN","plant","<self>"
"PUNCT",".","sil"
"<eos>","<eos>",""
"DATE","2006","two thousand six"
"LETTERS","IUCN","i u c n"
    \end{Verbatim}
    \caption{A sample from the original dataset.}
    \label{fig:dataset-orig}
\end{figure}

As shown in \hbox{Figure \ref{fig:dataset-orig}}, there are special symbols in the original dataset, namely: \hbox{(1) \eos,} denoting the end of the current sentence; \hbox{(2) \sil,} marking a silence (comma, colon, and so on); and \hbox{(3) \self,} meaning that the output in that entry is the same as the input. Since these symbols cannot be used in a character-level approach (due to the alignment problem), they are removed in the following way: \eos~disappears once the sentence has been recomposed; and the remaining symbols are substituted by the input, which is also the output.

Other minor changes have to made on the original dataset to speed up the training process, obtaining a dataset as shown in \hbox{Figure \ref{fig:dataset-final}}. The process\footnote{The code used for preprocessing the data is available at:  \url{https://github.com/adrianjav/text-normalization-preprocess}.} consists of the following steps:
\begin{enumerate}
    \item Concatenation of words belonging to the same phrase and removal of special symbols, as mentioned before.
    \item Phrases with non-permitted characters are discarded, keeping an alphabet of \hbox{$v = \num{127}$} characters, including numbers, simple arithmetic symbols, currency, and the English alphabet.
    \item Entries with an output longer than \num{177} characters are discarded as well, which represent only \SI{0.01}{\percent} of the population. 
    \item Entries are sorted in descending order with respect to their output length. This way, the padding introduced in batches is minimized and, as described by \citet{DBLP:conf/icml/XuBKCCSZB15}, convergence speed is increased without a significant loss in accuracy.
\end{enumerate}

\begin{figure}[hbtp]
    \centering
    \begin{Verbatim}[frame=single]
"Input Token","Output Token"
"Rosemary is a plant .","Rosemary is a plant ."
"2006 IUCN .","two thousand six i u c n ."
"We all lost .","We all lost ."
"vol 6 no","volume six no"
"Rees et al .","Rees et al ."
    \end{Verbatim}
    \caption{Sample entries from the preprocessed dataset.}
    \label{fig:dataset-final}
\end{figure}

\subsection{Experimental Setup}

After preprocessing the dataset, subsets of the final dataset have to be chosen in order to train and compare the models in a reasonable time. %
For this purpose, three experiments are prepared, each one having its own dataset. The first two datasets will be used to test and compare different models, whereas the latter will be used to train the final model and compare it with prior results. \hbox{Figure \ref{tab:experiments}} shows their names, training times, number of training elements, and the way entries have been selected: {\tt random} means that they have been randomly taken and {\tt shortest} that the elements with shortest outputs have been selected. In all cases, 1/5 additional entries are taken for the validation and for the test sets.

\begin{table}[hbtp]
	\centering
	\begin{tabular}{cccc}
		\hline
		{Name} & {Duration} & {Size} & {Selection} \\ \hline
		{E1} & \SI{1}{\hour} & \num{50000} & shortest \\
		{E2} & \SI{12}{\hour} & \num{50000} & random \\
		{E3} & \SI{22}{\hour} & \num{1000000} & random \\
		\hline
	\end{tabular}
	\caption{Description of the sets used in the experiments.}
	\label{tab:experiments}
\end{table}

Regarding the actual input and output used in the model, a one-hot encoding has been chosen, i.e., a string $s = s_0s_1\dots s_l$ of size $l\in\N$ will turn into a matrix $\X\in\M_{v\times l}$ whose $i$-th column $x_i\in\X$ is set to zero in every position but the one corresponding to the index of the character $s_i$ according to the model alphabet.

The advantages and disadvantages of using a character-level model have been described by other authors, since it appears as a design question in many NLP problems. Four arguments in favor of character-level approaches are shown, three of them introduced by \citet{DBLP:conf/acl/ChungCB16}, and the last one given by \citet{DBLP:journals/tacl/LeeCH17}:
\begin{itemize}
	\item Out-of-vocabulary issues do not appear anymore. We could suffer from out-of-alphabet issues, but these are easier to solve.
	\item Such approaches are able to model rare morphological variants of a word.
	\item Input segmentation is no longer required.
	\item By not segmenting, we encourage the models to discover the internal rules and structure of the sentences by themselves.
\end{itemize}

Since text segmentation is known to be problematic and error prone, even for well-known languages like English, getting rid of this step without losing performance is a significant advantage to take into account.

We present an additional argument for character-level approaches. If the model uses attention mechanisms, observing the attention matrices after a particular sample could allow us to gain a better understanding of the system's logic and the language itself. For example, consider the case where the model transforms \exampletext{2s} into \exampletext{two seconds}; its attention matrix could potentially show that the last letter was produced by looking at the number.

\subsection{Encoder-Decoder Architecture}
	
\begin{figure}[hbtp]
	\centering
		\includegraphics[keepaspectratio]{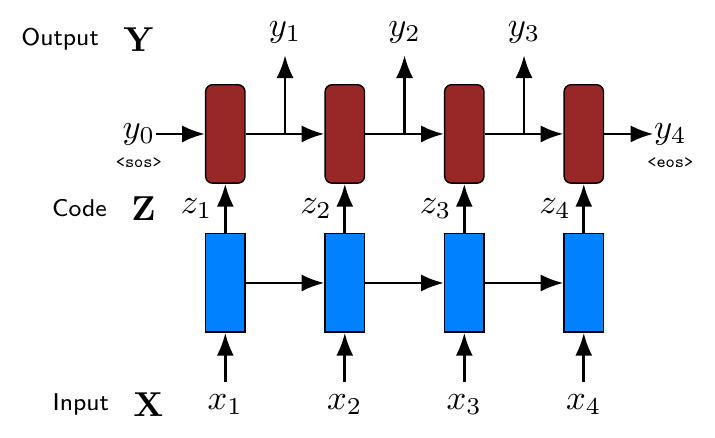}
		\caption{A basic encoder-decoder architecture. Blue: encoder. Red: decoder.}
		\label{fig:cod-dec-basic}
\end{figure}

The encoder-decoder architecture is a common design in recent {Neural Machine Translation} literature, and its architecture is easy to grasp. The model is composed of two parts: \hbox{(1) an} {encoder} that takes the input $\X$ (in this case, a phrase) and produces an intermediate representation $\Z$ (or {code}) that highlights its main features; and \hbox{(2) a} {decoder} that processes that set of features and produces the required output $\Y$ (in this case, a normalized phrase). \hbox{Figure \ref{fig:cod-dec-basic}} shows a basic diagram of this model.

There is a trend in using {Long Short-Term Memory} (LSTM) neural networks as encoders and decoders (for example, \citet{DBLP:conf/nips/SutskeverVL14}) due to their ability to capture long dependencies among the elements of a sequence. %
Our proposed model will use an LSTM network as decoder. However, different encoders will be analyzed, including the proposed one, and their performance will be tested and compared.


The basic encoder-decoder architecture looks great at first, but some key issues arise when they are put it on practice. Two of them stand out and are worth mentioning: \hbox{(1) as} shown in \hbox{Figure \ref{fig:cod-dec-basic}}, at each step the decoder works with the code produced at that moment, hindering the usage of long-term dependencies; and \hbox{(2) output} and input need to have the same length, constraining the suitable use cases of the model.

\begin{figure}[hbtp]
    \centering
    \includegraphics[keepaspectratio]{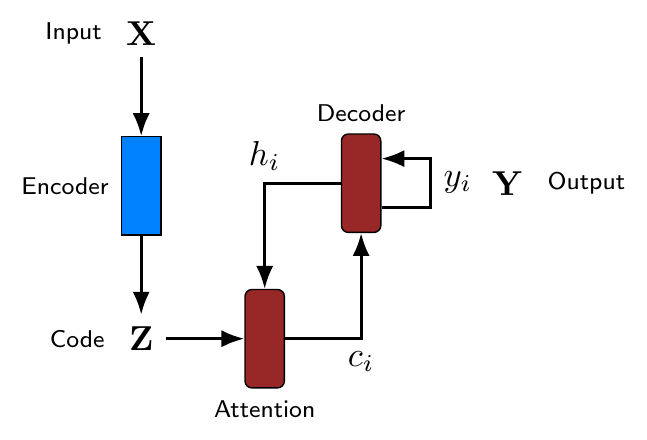}
    \caption{An encoder-decoder architecture with attention mechanisms.}
	\label{fig:cod-dec-attn}
\end{figure}

These two setbacks are overcome by the implementation of attention mechanisms \cite{DBLP:journals/corr/BahdanauCB14}. The idea behind them, depicted in \hbox{Figure \ref{fig:cod-dec-attn}}, is quite simple: first, produce the codes of the whole input sequence at once and, in each time step, let the decoder choose the most interesting elements of the input based on the latest output.

This can be expressed in mathematical terms as follows. Suppose that \hbox{$\Z = z_0z_1\dots z_k$} is the code at the $i$-th time step; then, the model gets a description of the interesting features $h_i$ (typically given by the decoder hidden states) and a small neural network produces a vector $\alpha = \alpha_0\alpha_1\dots\alpha_k$ from this description. This vector $\alpha$ is transformed into a stochastic vector, i.e., a vector such that $\sum \alpha_i = 1$, via
\[ 
\alpha_i = \frac{\exp \alpha_i}{\sum_j \exp \alpha_j}
\]
Now, $\alpha_i$ represents the interest of the decoder with respect to the $i$-th element of the code $z_i$ and a {context vector} is produced, that is, a vector representing the portion of the input that is actually interesting for the decoder. 

Traditionally, this context vector is taken as a weighted sum of the elements of $z_i$, weighted by $\alpha$, $c = \sum_i \alpha_i z_i$. A different approach is taken in this research. Instead of performing a weighted sum, a hyperparameter {\tt d} describing the number of context elements is set, and a {context matrix} $c$ is generated where the $i$-th column $c_i$ corresponds to the element $\alpha_i z_i$ having the $i$-th greatest value of $\alpha_i$, that is:
\[
c_i = a_jz_j \quad \text{where} \quad j = \argmax_{j=0}^k \left \{ a_j~\text{such that}~a_jz_j\neq c_l~\text{for}~0\leq l < i \right\}
\]

The idea inspiring this modification is that by not averaging the feature vectors, the internal semantic of each individual element is preserved.

\subsection{The Proposed Causal Feature Encoder} \label{subsubsec:cfe} 

\begin{figure*}[hbtp]
	\centering
	\begin{subfigure}{.49\textwidth}
    	\centering
		\includegraphics[keepaspectratio]{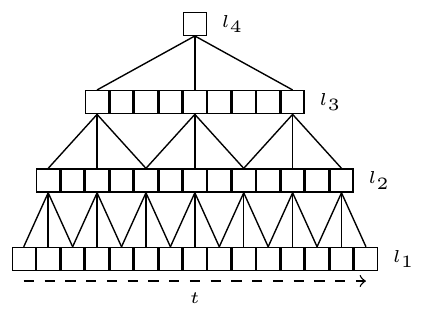}
		\caption{A regular CNN.}
		\label{fig:cnn}
	\end{subfigure} %
	\begin{subfigure}{.49\textwidth}
	    \centering
		\includegraphics[keepaspectratio]{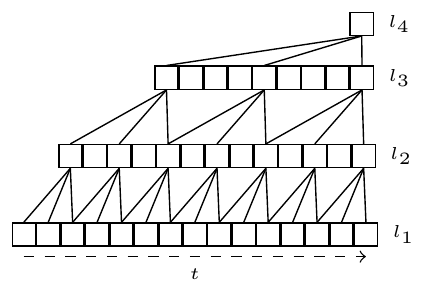}
		\caption{A causal CNN.}
		\label{fig:causal-cnn}
	\end{subfigure}
    \caption{A sample of regular (a) and casual (b) CNNs.}
\end{figure*}

The new encoder proposed in this paper can be described as a two-step modification of a traditional CNN. The first change is that, instead of using regular convolutions, {causal convolutions} (introduced by \citet{DBLP:conf/ssw/OordDZSVGKSK16}) are used. \hbox{Figures \ref{fig:cnn}} \hbox{and \ref{fig:causal-cnn}} show a basic representation of a regular and causal neural network, respectively.

\begin{figure}[hbtp]
    \centering
    \includegraphics[keepaspectratio]{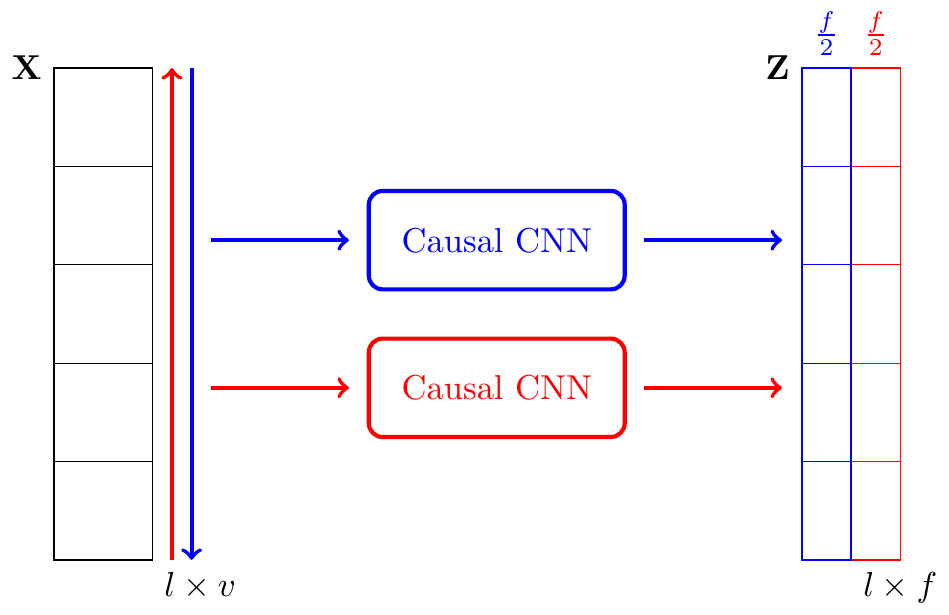}
    \caption{Diagram showing the bidirectionality of the proposed CFE.}
    \label{fig:cfe}
\end{figure}

The second step allows the CFE to solve an important drawback: it can only capture dependencies in one direction. To overcome it, the CFE is made {bidirectional} as with LSTMs. In this way, it contains two independent models that read the input in each direction and concatenate their outputs to produce the desired output. This is depicted in \hbox{Figure \ref{fig:cfe}}.

An additional change has been made. Because of the long sequences found in text normalization (up to \num{177} characters), the concept of {dilated convolutions} has been applied to CFE, as described by \citet{DBLP:conf/ssw/OordDZSVGKSK16}, doubling the dilatation of each layer as it goes deeper into the structure. By doing that, the actual receptive field of the model is significantly increased without increasing the number of parameters. 

This new encoder comes as an attempt to solve a problem that CNNs show with attention mechanisms. In previous experiments, it was observed that CNNs tend to attend the wrong inputs according to our prior intuition, namely they choose the \hbox{$i + C$-th} element instead of the $i$-th element, where $C$ is a constant. Our intuition is that this is caused by the padding introduced in each side. By using causal convolutions, the model is forced to choose the outermost elements if it is interested in those.

\subsection{Statistical Test}

When comparing various models, it is critical to ensure that the differences that can be appreciated are statistically significant. It has to be proved that those differences are actual differences, and not a product of the implicit variance coming from training the models. This is typically performed using some statistical test that will assert that the differences are actual differences up to some probability percentage, usually \SI{95}{\percent}.

For this article, we have opted for the {approximate randomization test} \cite{DBLP:conf/acl/RiezlerM05}. This statistical test measures the probability of the outputs of two different models of being indistinguishable, i.e., the probability that, by just looking at the predictions, we cannot tell whether those predictions come from different models. The main reasons for opting for this method are: \hbox{(1) it} is computationally cheap; \hbox{(2) it} is distribution-free, meaning that it does not make any assumptions on the distribution measured; and \hbox{(3) it} is model-free, that is, the only required resources to perform the test are the actual predictions, making it suitable for any kind of conceivable model.

Let us assume that the predictions are two ordered sets, $A$ and $B$, and that we have a function $e$ that measures the closeness of the predictions with respect to the actual solutions $Y$, for example, the accuracy. Then, we can define the function: 
\[
t(A, B) = | e(A, Y) - e(B, Y)|
\]
and we are seeking the probability of getting a bigger error than \hbox{$t(A, B)$}, assuming that both sets of predictions are indistinguishable, i.e., \hbox{$P(X \geq t(A, B) | H_0)$}.

\begin{figure}[hbtp]
	\centering
	\includegraphics[keepaspectratio, width=\linewidth]{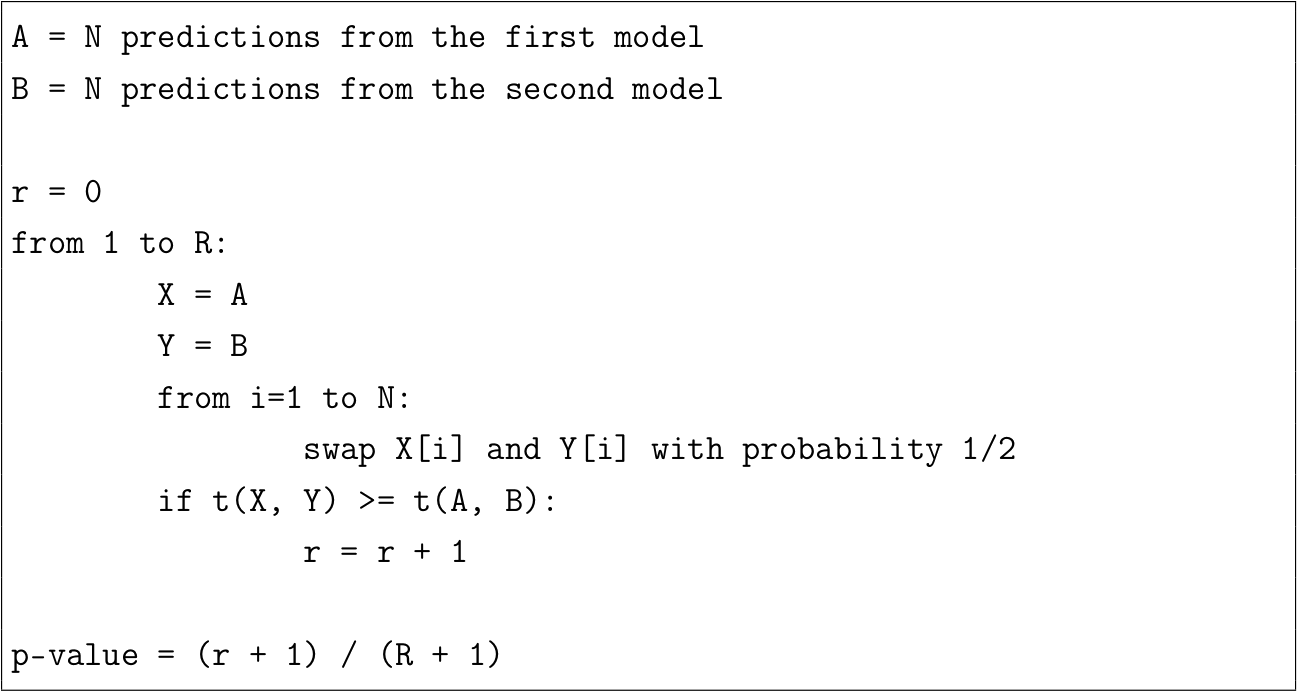}
	\caption{Pseudocode of the approximate randomization test. R is the number of repetitions selected.}
	\label{fig:pseudo}
\end{figure}

The algorithm that approximates this value just repeats many times (typically a thousand) the same process: it randomly swaps each element of the first set with its counterpart in the second set, and counts the number of times that the total error, measured by $t$, is greater or equal than the original one, that is, $t(A, B)$. \hbox{Figure \ref{fig:pseudo}} shows the pseudocode of this algorithm.

\section{Results}
\label{sec:results}

\subsection{Proposed Methods and Number of Parameters}

As said before, different encoders are considered to test whether CFE entails an actual improvement with respect other encoders. These encoders (and their alias) are the following:
\begin{itemize}[leftmargin=.15\linewidth]
	\item[LSTM] A simple bidirectional LSTM network.
	\item[FCNN] A FCNN encoder where the $i$-th element is an embedding of the $i$-th input.
	\item[FE] A traditional CNN with dilated convolutions.
	\item[CFE] The Causal Feature Extractor encoder.
\end{itemize}

Hyperparameters of each model were manually tuned, and the results have been averaged over five exact models trained with different random seeds.

The most basic question comparing multiple neural models concerns the number of trainable parameters. This data is quite easy to obtain, and knowing the numbers of parameters of a model, equivalently, its size ---and, to a lesser extent, its complexity--- can be a deciding point in case of a tie. The number of parameters of the models are shown in \hbox{Table \ref{tab:parameters}}.

\begin{table}[hbtp]
	\centering
	\begin{tabular}{lcccc}
		{Encoder} & {LSTM} & {FCNN} & {FE} & {CFE} \\
		{No. of parameters: encoder (millions)} & \num{1.102} & \num{0.285} & \num{0.111} & \num{0.111} \\
		{No. of parameters: total (millions)} & \num{7.380} & \num{6.653} & \num{6.479} & \num{6.479} \\
	\end{tabular}
	\caption{Number of parameters of each model.}
	\label{tab:parameters}
\end{table}

\subsection{First Experiment}
\label{subsec:first}

The results obtained for the first experiment are shown in \hbox{Table \ref{tab:results-e1}}. These results are averaged over five runs and extracted from the test set results, except from the results concerning the training speed, which are taken from the training logs. From left to right, the columns of \hbox{Figure \ref{tab:results-e1}} show:

\begin{itemize}
    \item {Negative Log-Likelihood Loss} (NLLLoss). It is the measure optimized by the neural network during training, since it is the usual measure in a classification setting.
    
    \item {Character Error Rate} (CER). It is defined as the Levenshtein distance between the prediction and the expected value, measured in characters.
    
    \item {Accuracy}. It is a basic and well-known measure, defined as the percentage of correct predictions.
    
    \item {Number of iterations} performed during the training phase in the duration of the experiment (in this case 1 hour).
    
    \item {Rate}. Number of iterations per second, on average, achieved during training.
\end{itemize}

\begin{table}[hbtp]
	\centering
	\begin{tabular}{@{\extracolsep{4pt}}cccccc@{}}
	\cline{2-4} \cline{5-6} 
	& \multicolumn{3}{c}{Test} & \multicolumn{2}{c}{Validation} \\ \cline{2-4} \cline{5-6} \\[-3mm]\hline
	{Encoder} & {NLLLoss} & {CER (\%)} & {Acc (\%)} & {No. iters} & {Rate} \\ \hline
	{LSTM} & \num{1.352} & \num{03.13} & \num{95.87} & \num{3620} & \num{1.005} \\
	{FCNN} & \num{5.035} & \num{70.61} & \num{28.40} & \num{4370} & \num{1.214} \\
	{FE} & \num{1.042} & \num{02.52} & \num{96.46} & \num{6980} & \num{1.939} \\
	{CFE} & \num{0.952} & \num{02.24} & \num{96.83} & \num{6300} & \num{1.750} \\
	\hline
	\end{tabular}
	\caption{Results obtained for the first experiment (E1).}
 	\label{tab:results-e1}
\end{table}

\begin{figure}[hbtp]
    \centering
    \includegraphics[keepaspectratio, width=\linewidth]{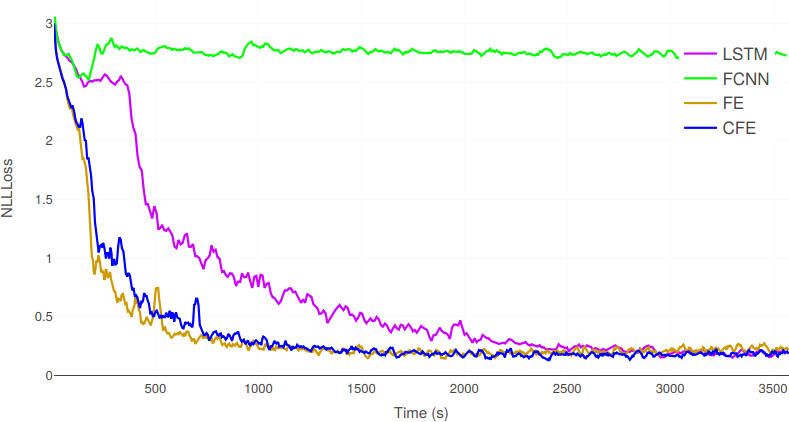}
    \caption{Evolution of the validation error during training on E1.}
    \label{fig:plot-e1}
\end{figure}

In order to get an understanding of the differences in the training process, \hbox{Figure \ref{fig:plot-e1}} shows the evolution of the NLLLoss over the validation tests of each model during the training process. Table \ref{tab:test-e1} shows the resulting p-values after running the approximate randomization test over each pair of models.

\begin{table}[hbtp]
    \centering
    	\begin{tabular}{rcccc}
		{p-value} & {LSTM} & {FCNN} & {FE} & {CFE} \\ \cline{2-5}
		\multicolumn{1}{r|}{{LSTM}} & \multicolumn{1}{c|}{} & \multicolumn{1}{c|}{\num{0.001}} & \multicolumn{1}{c|}{\num{0.001}} & \multicolumn{1}{c|}{\num{0.003}} \\ \cline{2-5}
		\multicolumn{1}{r|}{{FCNN}} & \multicolumn{1}{c|}{\num{0.001}} & \multicolumn{1}{c|}{} & \multicolumn{1}{c|}{\num{0.001}} & \multicolumn{1}{c|}{\num{0.001}} \\ \cline{2-5}
		\multicolumn{1}{r|}{{FE}} & \multicolumn{1}{c|}{\num{0.001}} & \multicolumn{1}{c|}{\num{0.001}} & \multicolumn{1}{c|}{} & \multicolumn{1}{c|}{\num{0.019}} \\ \cline{2-5}
		\multicolumn{1}{r|}{{CFE}} & \multicolumn{1}{c|}{\num{0.003}} & \multicolumn{1}{c|}{\num{0.001}} & \multicolumn{1}{c|}{\num{0.019}} & \multicolumn{1}{c|}{} \\ \cline{2-5}
	\end{tabular}
    \caption{P-values of the first experiment (E1).}
    \label{tab:test-e1}
\end{table}

\subsection{Second Experiment}

\begin{figure}
    \centering
    \includegraphics[keepaspectratio, width=\linewidth]{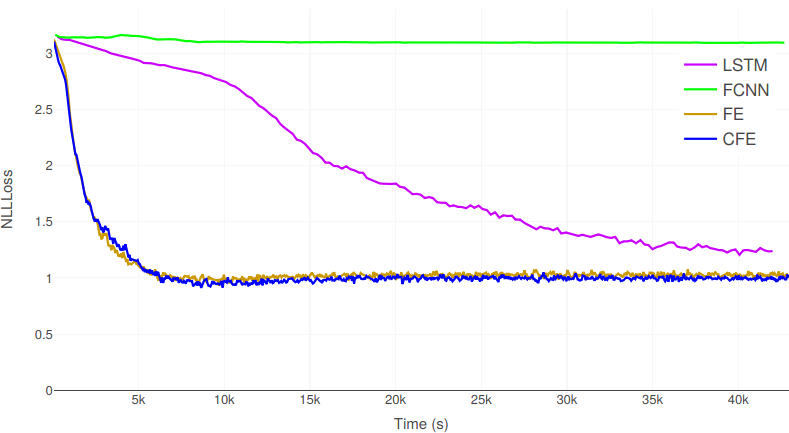}
    \caption{Evolution of the validation error during training on E2.}
    \label{fig:plot-e2}
\end{figure}

As before, \hbox{Table \ref{tab:results-e2}} shows the same measures, but now regarding the second experiment. \hbox{Figure \ref{fig:plot-e2}} and \hbox{Table \ref{tab:test-e2}} show the evolution of the validation error and the results of the statistical test on the second experiment, respectively.

\begin{table}[hbtp]
	\centering
	\begin{tabular}{@{\extracolsep{4pt}}cccccc@{}}
	\cline{2-4} \cline{5-6} 
	& \multicolumn{3}{c}{Test} & \multicolumn{2}{c}{Validation} \\ \cline{2-4} \cline{5-6} \\[-3mm]\hline
	{Encoder} & {NLLLoss} & {CER (\%)} & {Acc (\%)} & {No. iters} & {Rate} \\ \hline
	{LSTM} & \num{3.310} & \num{25.59} & \num{71.06} & \num{8900} & \num{0.206} \\ 
	{FCNN} & \num{5.396} & \num{82.09} & \num{17.90} & \num{17750} & \num{0.411} \\
	{FE} & \num{2.680} & \num{11.93} & \num{83.38} & \num{36650} & \num{0.848} \\
	{CFE} & \num{2.686} & \num{12.69} & \num{83.45} & \num{36200} & \num{0.838} \\
	\hline
	\end{tabular}
	\caption{Results obtained for the second experiment (E2).}
 	\label{tab:results-e2}
\end{table}

\begin{table}[hbtp]
    \centering
    	\begin{tabular}{rcccc}
		{p-value} & {LSTM} & {FCNN} & {FE} & {CFE} \\ \cline{2-5}
		\multicolumn{1}{r|}{{LSTM}} & \multicolumn{1}{c|}{} & \multicolumn{1}{c|}{\num{0.001}} & \multicolumn{1}{c|}{\num{0.001}} & \multicolumn{1}{c|}{\num{0.001}} \\ \cline{2-5}
		\multicolumn{1}{r|}{{FCNN}} & \multicolumn{1}{c|}{\num{0.001}} & \multicolumn{1}{c|}{} & \multicolumn{1}{c|}{\num{0.001}} & \multicolumn{1}{c|}{\num{0.001}} \\ \cline{2-5}
		\multicolumn{1}{r|}{{FE}} & \multicolumn{1}{c|}{\num{0.001}} & \multicolumn{1}{c|}{\num{0.001}} & \multicolumn{1}{c|}{} & \multicolumn{1}{c|}{\num{0.001}} \\ \cline{2-5}
		\multicolumn{1}{r|}{{CFE}} & \multicolumn{1}{c|}{\num{0.001}} & \multicolumn{1}{c|}{\num{0.001}} & \multicolumn{1}{c|}{\num{0.001}} & \multicolumn{1}{c|}{} \\ \cline{2-5}
	\end{tabular}
    \caption{P-values for the second experiment (E2).}
    \label{tab:test-e2}
\end{table}

\subsection{Third Experiment}

\begin{figure}[hbtp]
    \centering
    \includegraphics[keepaspectratio, width=\linewidth]{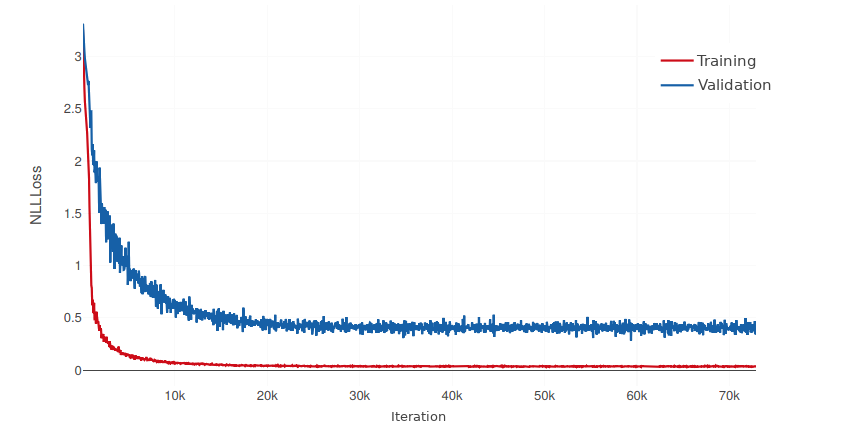}
    \caption{Evolution of the training (red) and validation (blue) errors over E3.} 
    \label{fig:plot-e3}
\end{figure}

In this subsection, the results of the final model after running the third experiment are shown. The final architecture is identical to the one with the CFE encoder used in the previous experiments. \hbox{Figure \ref{fig:plot-e3}} depicts the evolution of the training and validation error during the training phase, and \hbox{Table \ref{tab:results-e3}} shows the results obtained for the test set.

\begin{table}[hbtp]
	\centering
	\begin{tabular}{cccc}
	\cline{2-4}
	& \multicolumn{3}{c}{Test} \\ \cline{2-4} \\[-3mm]\hline
	{Encoder} & {NLLLoss} & {CER (\%)} & {Acc (\%)} \\ \hline
	{CFE} & \num{1.701} & \num{5.44} & \num{92.74} \\
	\hline
	\end{tabular}
	\caption{Results on the test set for the third experiment (E3).}
 	\label{tab:results-e3}
\end{table}

\subsection{Attention Matrices}

In order to show whether the CFE encoder makes a better usage of the attention mechanisms than its non-causal counterpart, it is necessary to show some actual examples and the attention matrices that they generate. These matrices are a representation of the decoder focus while it was processing the input: the \hbox{$i$-th} row represents the \hbox{$i$-th} character it predicted, and the \hbox{$j$-th} column is the model focus while predicting that character.

\begin{figure*}[hbtp]
	\centering
	\begin{subfigure}{.49\textwidth}
    	\centering
        \includegraphics[keepaspectratio, width=\linewidth]{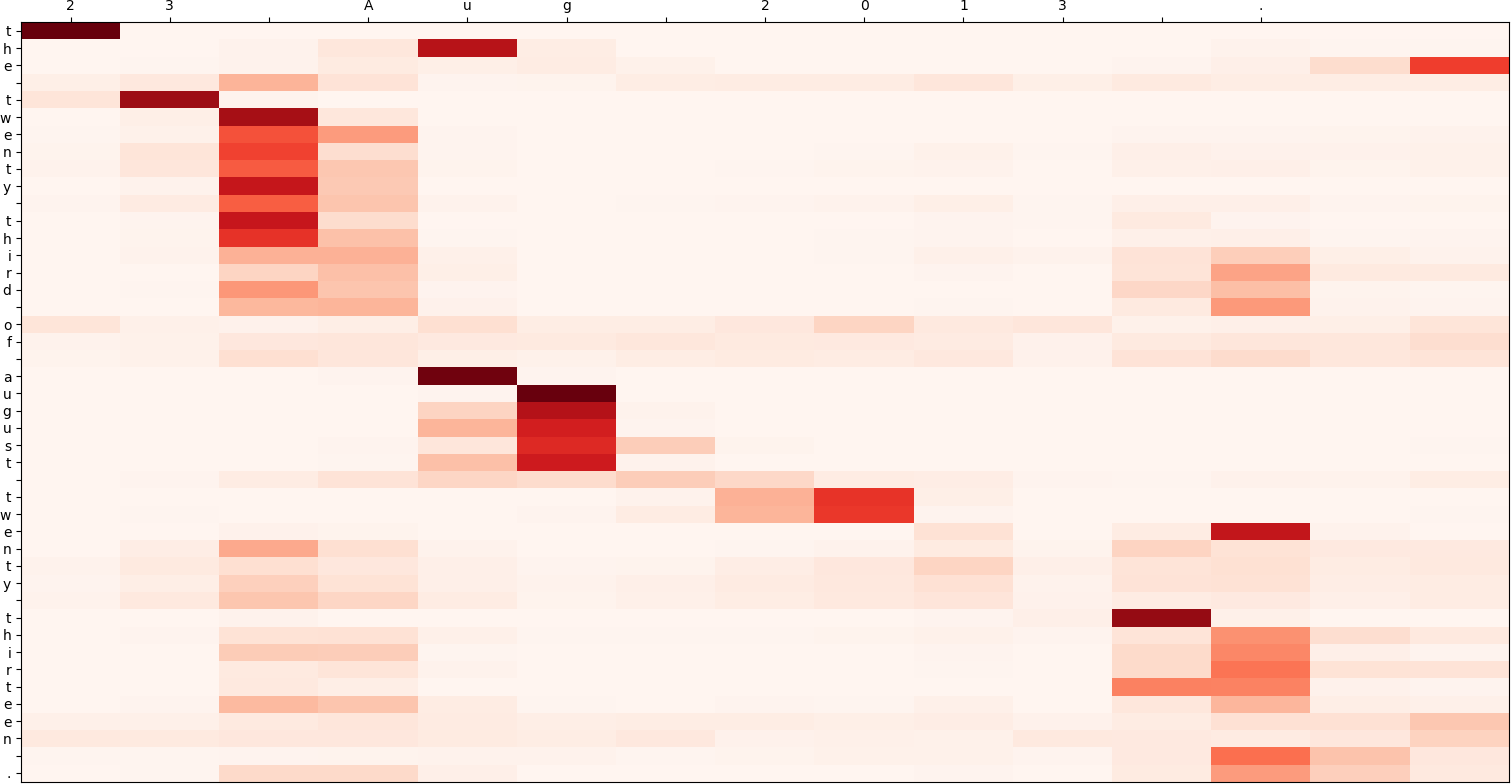}
        \caption{LSTM}
	\end{subfigure} %
	\begin{subfigure}{.49\textwidth}
	    \centering
        \includegraphics[keepaspectratio, width=\linewidth]{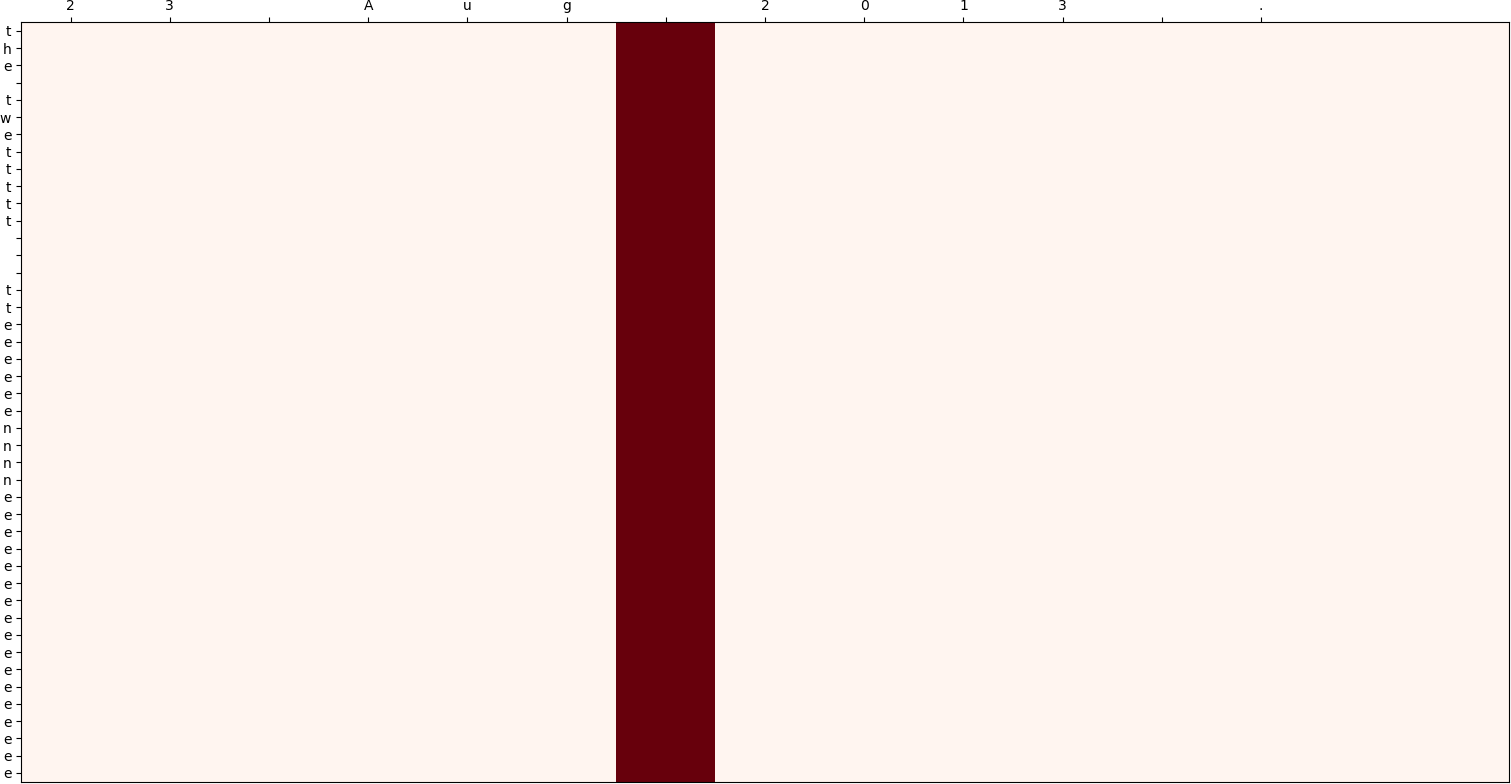}
        \caption{FCNN}
	\end{subfigure}
	\begin{subfigure}{.49\textwidth}
    	\centering
        \includegraphics[keepaspectratio, width=\linewidth]{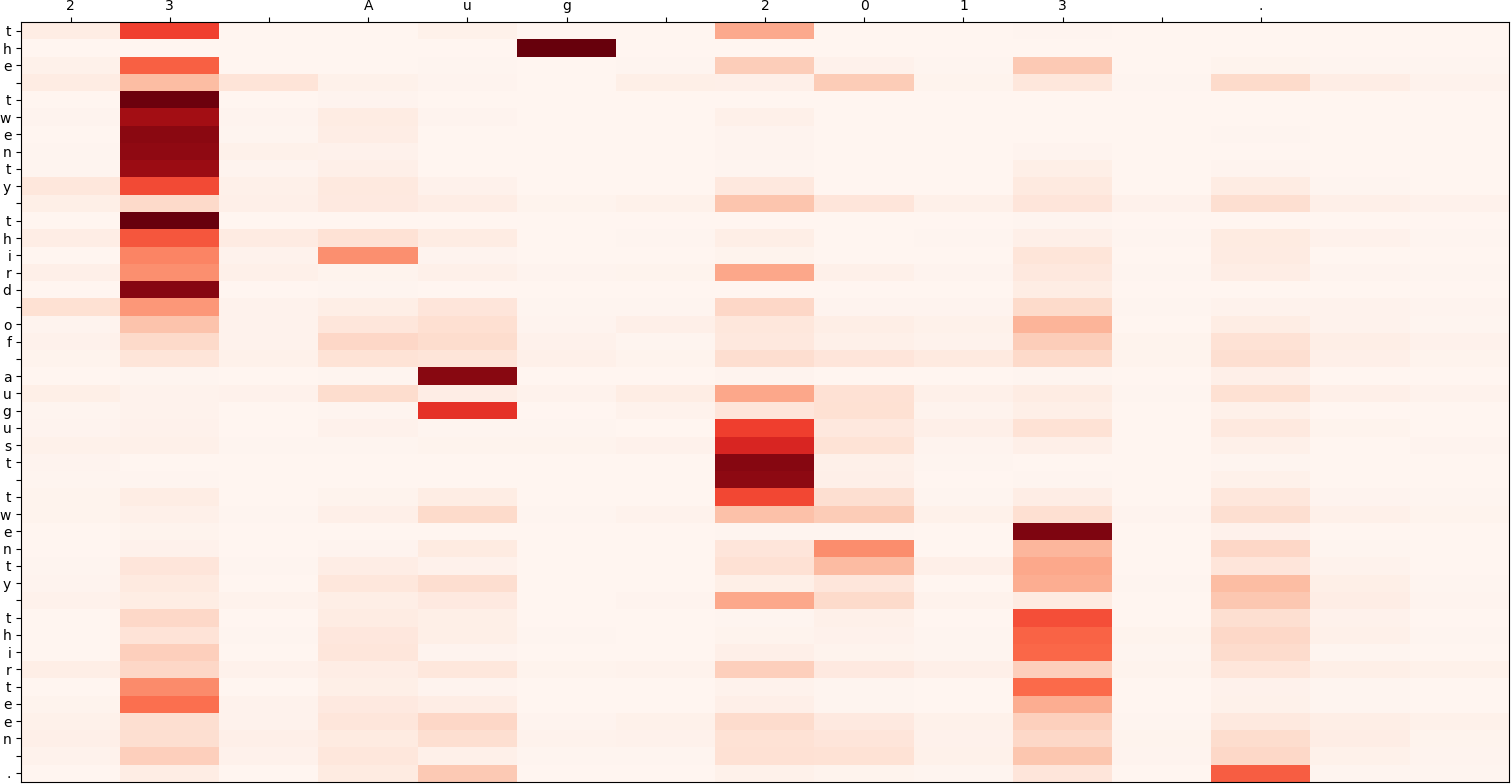}
        \caption{FE}
	\end{subfigure} %
	\begin{subfigure}{.49\textwidth}
	    \centering
        \includegraphics[keepaspectratio, width=\linewidth]{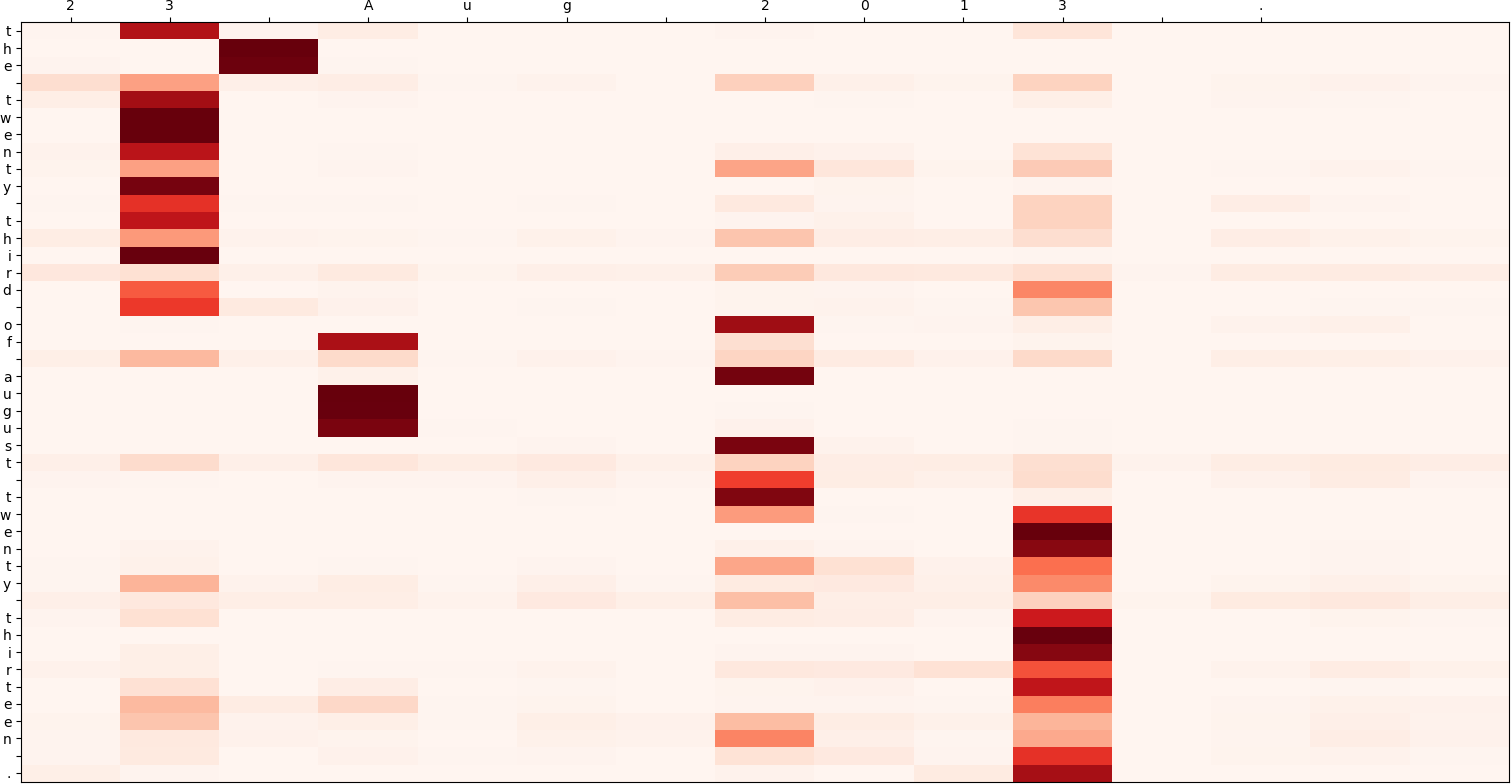}
        \caption{CFE}
	\end{subfigure}
    \caption{Attention matrices of an example of E1.}
    \label{fig:attn-mat-e1}
\end{figure*}

The first case, shown in \hbox{Table \ref{tab:example-e1}}, is an example extracted from the test set of the first experiment. The input phrase is \exampletext{23\space Aug\space 2013}. Regarding what it would be expected from the attention matrix to look like, it can expressed in three phases: \hbox{(1) it} writes out the day while focusing on its digits; \hbox{(2) shifts} its attention towards the month; and \hbox{(3) it} finishes by looking at the year. \hbox{Figure \ref{fig:attn-mat-e1}} shows the attention matrices.

\begin{table}[!hbtp]
    \centering
    \begin{tabular}{lcl}
         {Input} &  & \exampletext{23\space Aug\space 2013\space .} \\
         {Output} & & \exampletext{the\space twenty\space third\space of\space august\space twenty\space thirteen\space .} \\
         {LSTM} & \cmark & \exampletext{the\space twenty\space third\space of\space august\space twenty\space thirteen\space .} \\
         {FCNN} & \xmark & \exampletext{the\space twent\space \space t\space \space \space t\space \space \space eeeeeeeeeeeeeeeeeeeeee}\dots \\
         {FE} & \cmark & \exampletext{the\space twenty\space third\space of\space august\space twenty\space thirteen\space .} \\
         {CFE} & \cmark & \exampletext{the\space twenty\space third\space of\space august\space twenty\space thirteen\space .}
    \end{tabular}
    \caption{Predictions of the different models for the first example.}
    \label{tab:example-e1}
\end{table}

\begin{figure*}[hbtp]
	\centering
	\begin{subfigure}{.49\textwidth}
    	\centering
        \includegraphics[keepaspectratio, width=\linewidth]{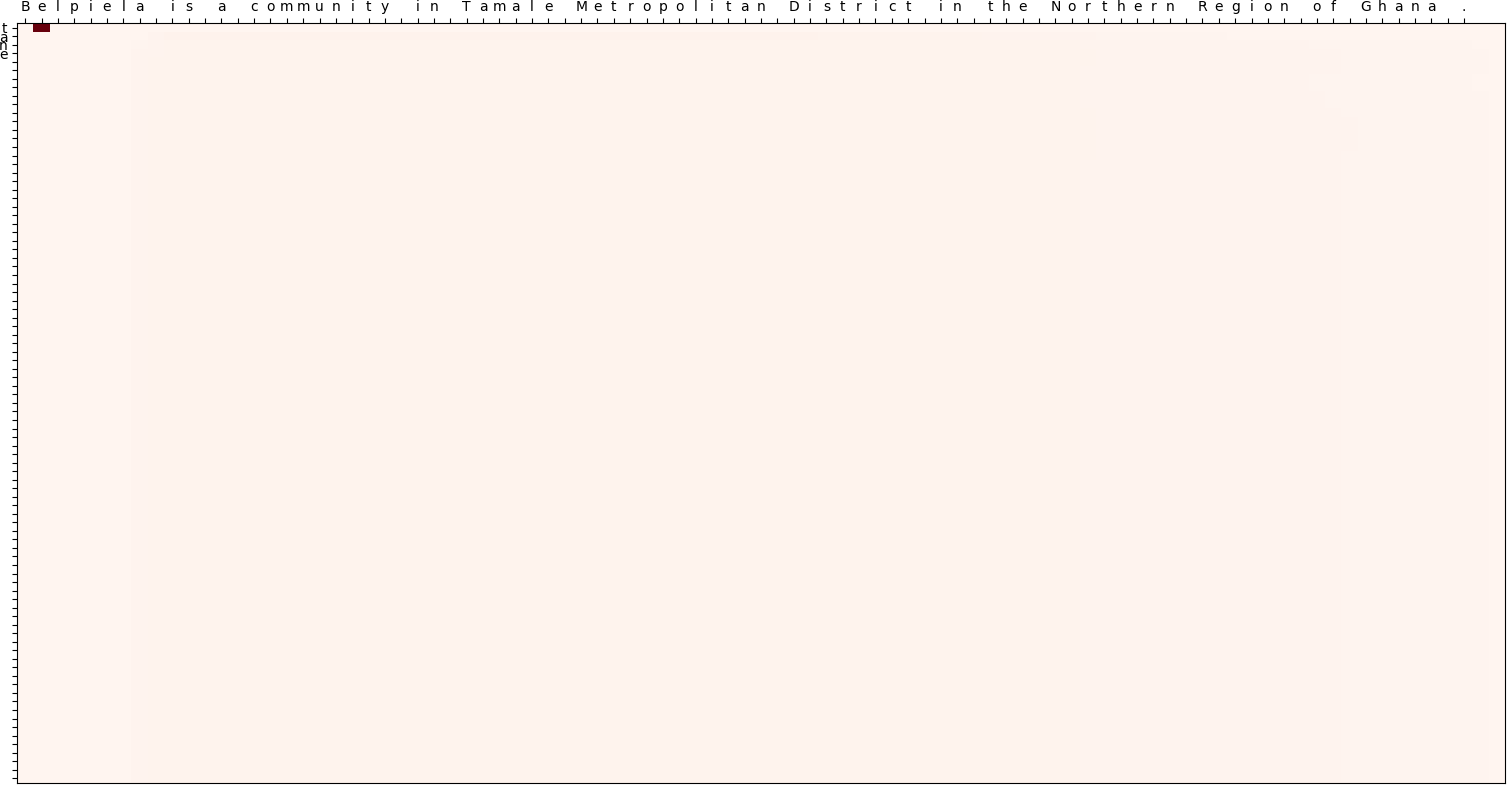}
        \caption{LSTM}
	\end{subfigure} %
	\begin{subfigure}{.49\textwidth}
	    \centering
        \includegraphics[keepaspectratio, width=\linewidth]{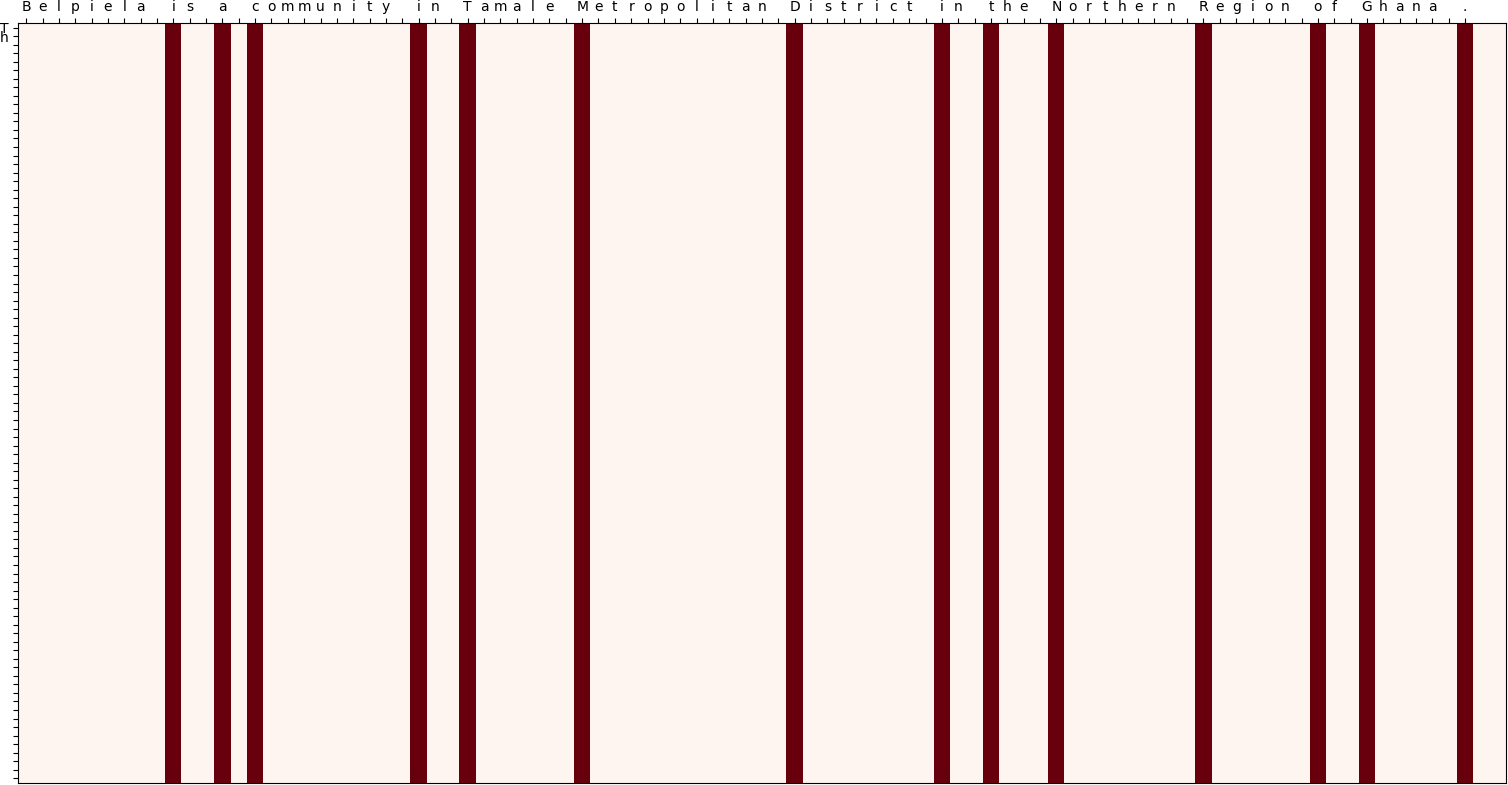}
        \caption{FCNN}
	\end{subfigure}
	\begin{subfigure}{.49\textwidth}
    	\centering
        \includegraphics[keepaspectratio, width=\linewidth]{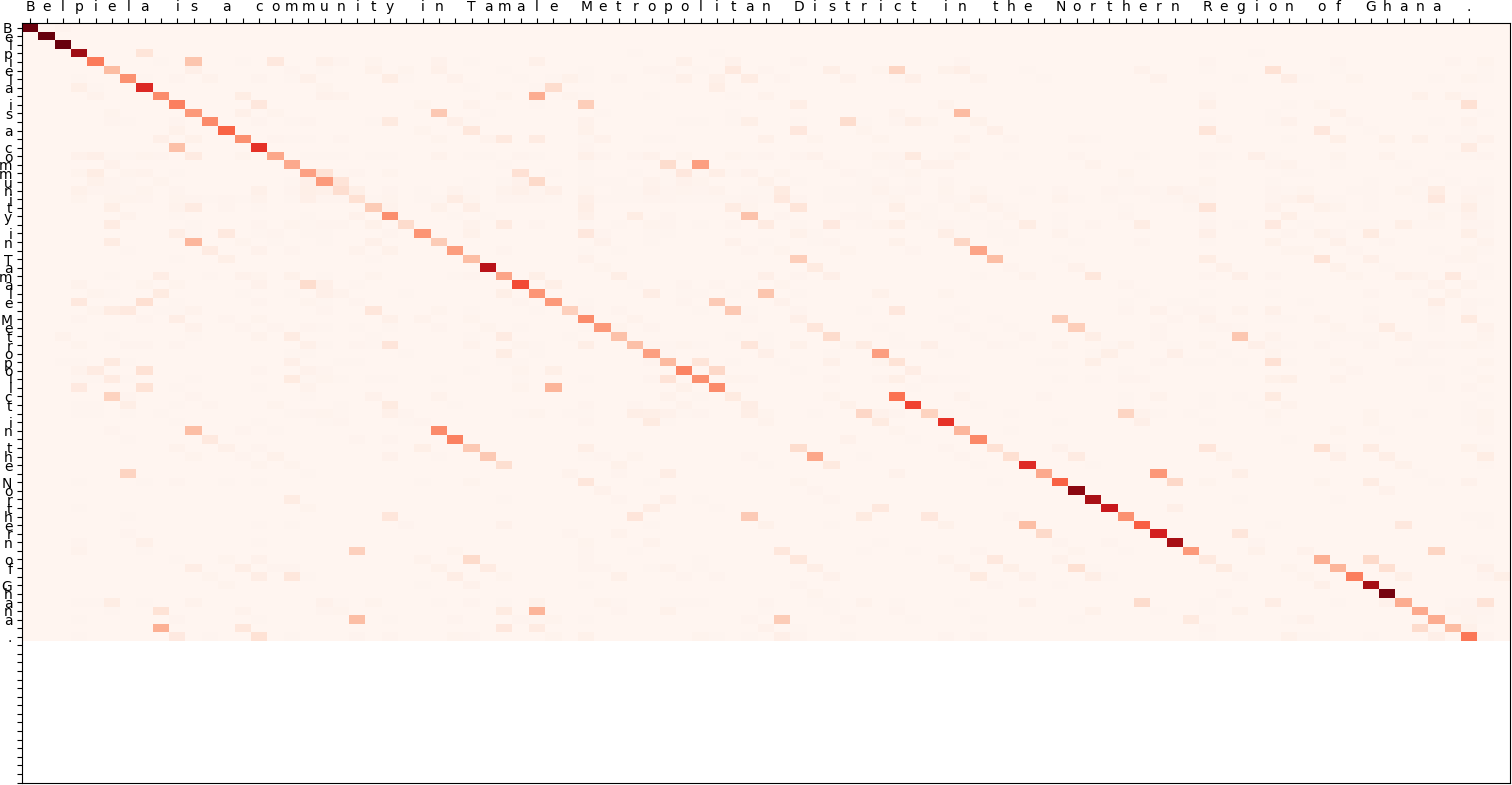}
        \caption{FE}
	\end{subfigure} %
	\begin{subfigure}{.49\textwidth}
	    \centering
        \includegraphics[keepaspectratio, width=\linewidth]{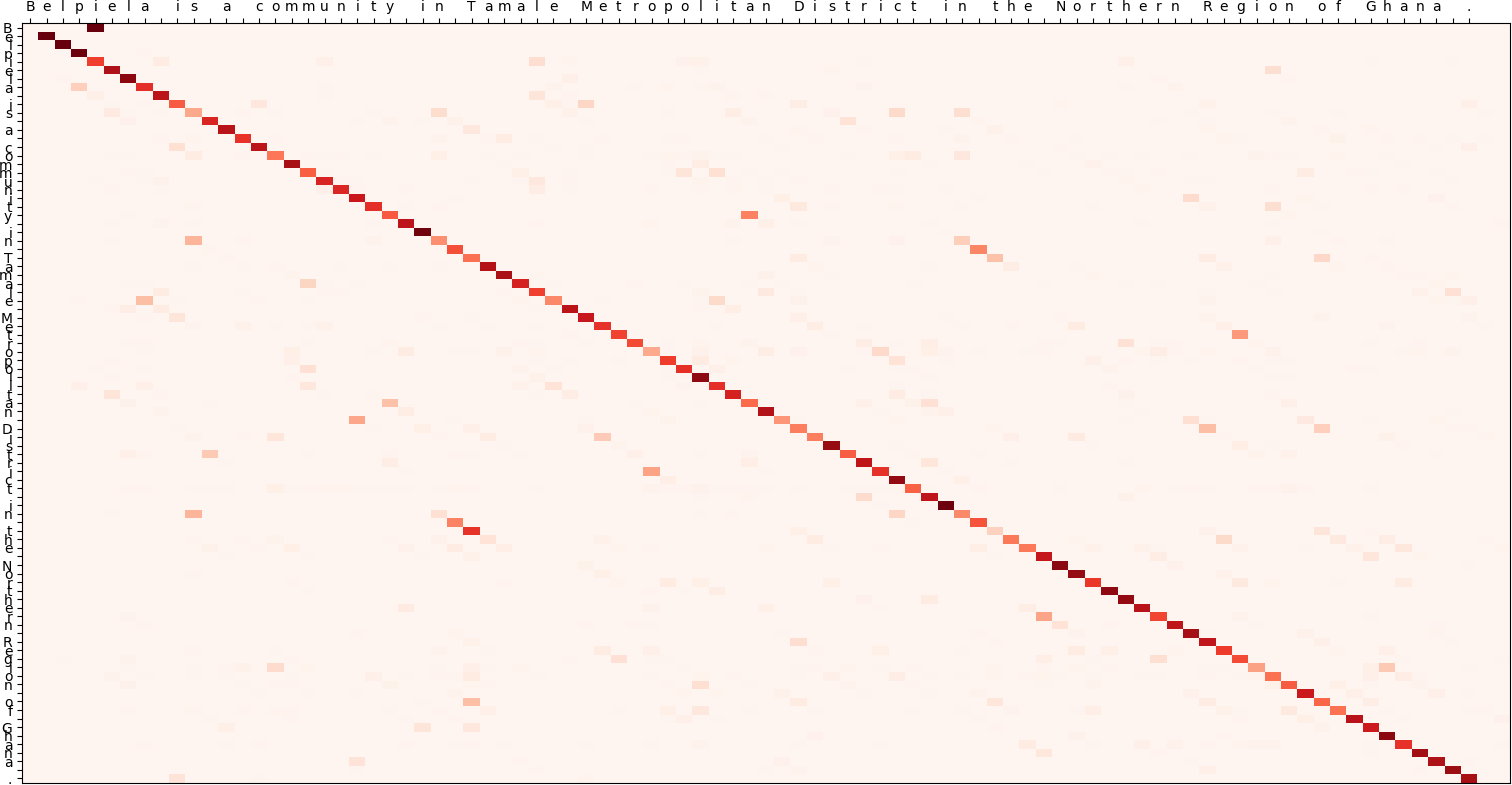}
        \caption{CFE}
	\end{subfigure}
    \caption{Attention matrices of an example of E2.}
    \label{fig:attn-mat-e2}
\end{figure*}

The second example has been taken from the test set of the second experiment. \hbox{Table \ref{tab:example-e2}} shows the predictions, whereas \hbox{Figure \ref{fig:attn-mat-e2}} shows the attention matrices. This is an example where the input and output are identical, and so, the expected attention matrices should resemble an identity matrix.

\begin{table}[!hbtp]
    \centering
    \begin{tabular}{lcl}
         {Input} &  & \exampletext{Belpiela\space is\space a\space community\space in\space Tamale\space Metropolitan\space District\space in\space the} \\ 
         & & \exampletext{Northern\space Region\space of\space Ghana\space .} \\
         {Output} & & \exampletext{Belpiela\space is\space a\space community\space in\space Tamale\space Metropolitan\space District\space in\space the} \\ 
         & & \exampletext{Northern\space Region\space of\space Ghana\space .} \\
         {LSTM} & \xmark & \exampletext{Belpiela\space is\space a\space community\space in\space Tamale\space Metropolitan\space Disire\space \space } \\
         & & \exampletext{egion\space te\space i\space \space \space \space e\space \space \space \qquad\qquad\qquad\qquad}\dots \\
         {FCNN} & \xmark & \exampletext{"\qquad\qquad\qquad\qquad\qquad}\dots \\
         {FE} & \xmark & \exampletext{Belpiela\space is\space a\space community\space in\space Tamale\space Metropolitan\space District} \\ 
         & & \exampletext{in\space the\space Northern\space Region\space Region\space of\space Ghana\space .} \\
         {CFE} & \cmark & \exampletext{Belpiela\space is\space a\space community\space in\space Tamale\space Metropolitan} \\ 
         & & \exampletext{District\space in\space the\space Northern\space Region\space of\space Ghana\space .}
    \end{tabular}
    \caption{Predictions of the different models for the second example.}
    \label{tab:example-e2}
\end{table}

\subsection{Error analysis}

In order to get a better understanding of the type of errors of the proposed model, those produced on the test set by the model of the third experiment has been dumped and analyzed by hand. Based on these observations, the taxonomy of the errors has been defined as follows:

\begin{enumerate}[label={T\theenumi}]
    \item Infinite loop errors. The attention system of the model gets stuck and the maximum number of printed characters is reached. For example:
        \begin{itemize}[leftmargin=6em]
            \item[Input] \exampletext{Ruppert , Edward E. ; Fox , Richard , S. ; Barnes , Robert D. ( 2004 ) .}
            \item[Output] \exampletext{Ruppert , Edward e ; Fox , Richard , s ; Barnes , Robert d ( two thousand four ) .}
            \item[Prediction] \exampletext{Ruppert , Edward e ; Fox , Richard , s R s , , , , , , , , , , , , , , , , , , ,} \dots
        \end{itemize}
    \item Coincidental errors. Predictions where only a few isolated characters are wrongly printed. For example:
        \begin{itemize}[leftmargin=6em]
            \item[Input] \exampletext{The income was \$11,091 .}
            \item[Output] \exampletext{The income was eleven thousand ninety one dollars .}
            \item[Prediction] \exampletext{The income was fleven thousand ninety one dollars .}
        \end{itemize}
    \item Early stop errors. Errors where the model finishes before processing the whole input.  For example:
        \begin{itemize}[leftmargin=6em]
            \item[Input] \exampletext{Parmentier , Bruno ( 2000-05-01 ) .}
            \item[Output] \exampletext{Parmentier , Bruno ( the first of may two thousand ) .}
            \item[Prediction] \exampletext{Parmentier , Bruno ( .}
        \end{itemize}
    \item (Finite) jumps. The attention model finds the same pattern in the entry and repeats/oversees part of it. For example:
        \begin{itemize}[leftmargin=6em]
            \item[Input] \exampletext{According to the 2011 census of India , Bhisenagar has 818 households .}
            \item[Output] \exampletext{According to the twenty eleven census of India , Bhisenagar has eight hundred eighteen households .}
            \item[Prediction] \exampletext{According to the twenty eleven census of India , Bhisenagar has eighteen households .}
        \end{itemize}
\end{enumerate}

A simple classification tool has been implemented in order to (approximately) quantify the errors according to their type. Results are shown in Table \ref{tab:errors} where {\em Others} refers to errors unclassified by the tool. Note that errors produced by jumps are not detected by the tool, but represent a big portion of the unclassified errors.

\begin{table}[hbtp]
	\centering
	\begin{tabular}{cccccc}
		\hline
		{Type} & {T1} & {T2} & {T3} & {Others} & {Total} \\ \hline
		{Quantity} & \num{23381} & \num{7159} & \num{50} & \num{10696} & \num{41286} \\
		{Percentage} (\SI{}{\percent}) & \num{56.63} & \num{17.34} & \num{0.12} & \num{25.9} & \num{100} \\ 
		\hline
	\end{tabular}
	\caption{Errors distribution from the test set of E3.}
	\label{tab:errors}
\end{table}

Besides these types of errors, it is worth-mentioning those caused by the dataset itself. These come from different sources, for example, from inconsistent rules for normalizing text among different entries, e.g.,
\begin{itemize}[leftmargin=6em]
    \item[Input] \exampletext{Uppsala : Sprak och folkminnesinstitutet ( SOFI ) .}
    \item[Output] \exampletext{Uppsala : Sprak och folkminnesinstitutet ( SOFI ) .}
    \item[Prediction] \exampletext{Uppsala : Sprak och folkminnesinstitutet ( S o f i ) .}
\end{itemize}
\begin{itemize}[leftmargin=6em]
    \item[Input] \exampletext{Chloroformic acid has the formula ClCO 2 H .}
    \item[Output] \exampletext{Chloroformic acid has the formula c l c o two H .}
    \item[Prediction] \exampletext{Chloroformic acid has the formula ClCO two H .}
\end{itemize}
by providing a few entries for rare cases that resemble too much to others, e.g.,
\begin{itemize}[leftmargin=6em]
    \item[Input] \exampletext{1980 A engine added to Transporter ( T 3 ) .}
    \item[Output] \exampletext{one nine eight o A engine added to Transporter ( T three ) .}
    \item[Prediction] \exampletext{nineteen eighty A engine added to Transporter ( T three ) .}
\end{itemize}
by inconsistency in the entries (e.g. American vs British) text, e.g.,
\begin{itemize}[leftmargin=6em]
    \item[Input] \exampletext{The mobilisation was announced by the mayor .}
    \item[Output] \exampletext{The mobilization was announced by the mayor .}
    \item[Prediction] \exampletext{The mobilisation was announced by the mayor .}
\end{itemize}
\begin{itemize}[leftmargin=6em]
    \item[Input] \exampletext{The Robinsons are a family in the soap opera Neighbours .}
    \item[Output] \exampletext{The Robinsons are a family in the soap opera neighbors .}
    \item[Prediction] \exampletext{The Robinsons are a family in the soapera Neighbors .}
\end{itemize}

\begin{figure}[hbtp]
    \centering
    \includegraphics[width=\textwidth]{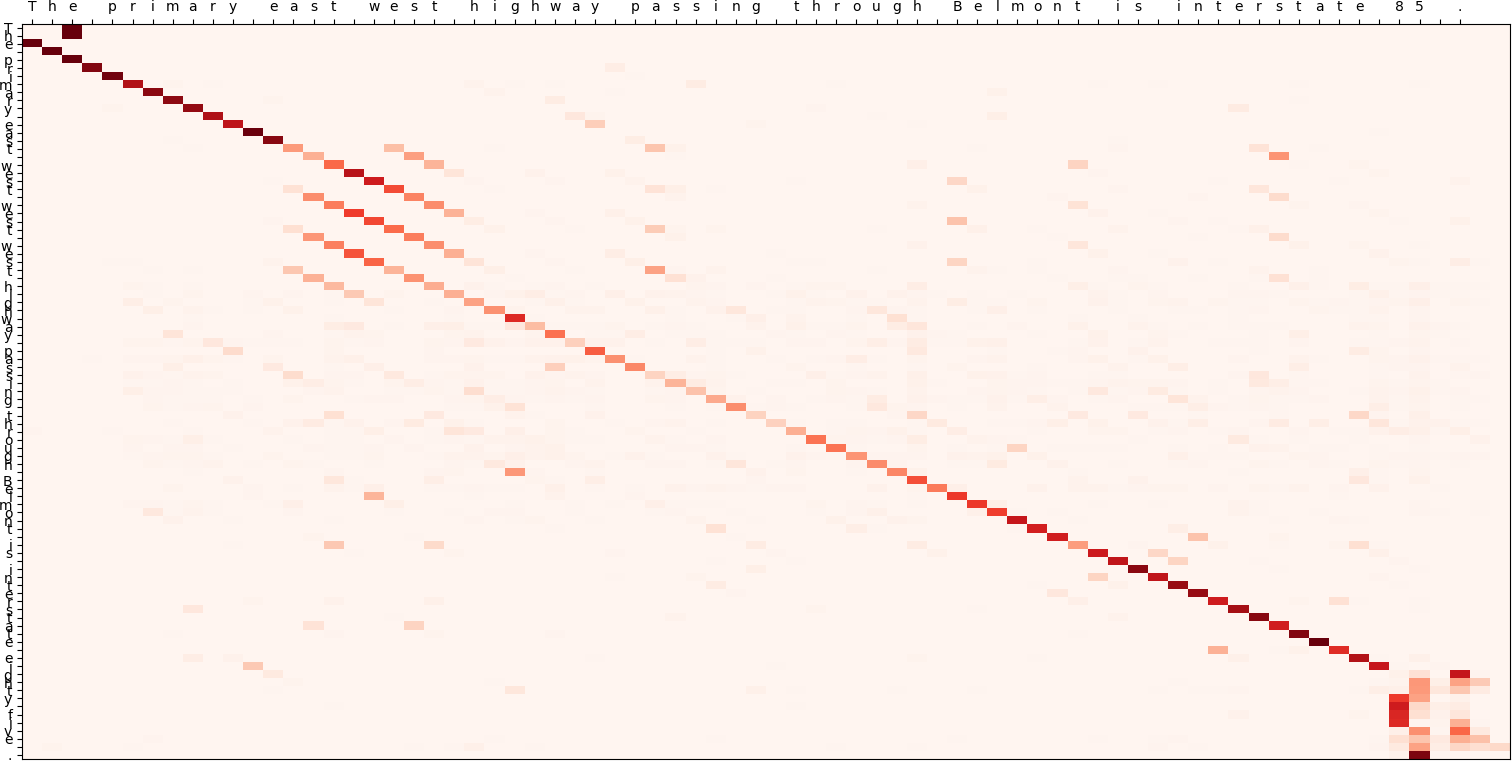}
    \caption{Attention matrix for a backwards finite jump error.}
    \label{fig:mat-error}
\end{figure}

The model specially struggles deciding whether it should maintain capital letters on the predictions. This last example also contains an example of overseeing parts of the input, probably because the similarities between the words \exampletext{soap} and \exampletext{opera}. Another example of such jumps, in this case going backward in the input and thus repeating words, is the following:
\begin{itemize}[leftmargin=6em]
    \item[Input] \exampletext{The primary east west highway passing through Belmont is interstate 85 .}
    \item[Output] \exampletext{The primary east west highway passing through Belmont is interstate eighty five .}
    \item[Prediction] \exampletext{The primary east west west west highway passing through Belmont is interstate eighty five .}
\end{itemize}
which happened because the model confounds the suffix of \exampletext{west} with the one of \exampletext{east} as it can be seen on Figure \ref{fig:mat-error}. Attention matrices can be displayed for all these errors shedding light on the attention-related issue underlying, except for the coincidental errors. 

Finally, is it worth-noting the role of undetectable errors as they were a problem in prior work. Among the analyzed test errors it can be found some like those, such as the one shown in the example for error T4. However, in all cases they are a realization of another type of error that happens to look like that by chance. The errors source can be explained and solved more readily and patterns in attention matrices could be leveraged to detect them. For example, the aforementioned error occurs as a realization of a jump error where the model confounds in \exampletext{818} the first \exampletext{8} with the third one when processing the input.

\section{Discussion}
\label{sec:discussion}

This section discusses the results presented in \hbox{Section \ref{sec:results}}. Specifically, the main questions stated in \hbox{Section \ref{subsec:goals}} were:
\begin{enumerate}
    \item Can the problem of text normalization be solved just by making use of neural networks?
    \item Is it viable such a solution using convolutional components? And which decoder is better?
\end{enumerate}

Answering the first question, the most obvious result we can extract, based on any of the results from E1 and E2 (for example, \hbox{Figure \ref{fig:plot-e2}}), is that the FCNN encoder does not work at all. Most probably, this erratic behavior comes from the differentiating feature of FCNN, that is, it extracts information of a single character of the input (instead of a neighborhood of it). This serves as a proof of an unsurprising result: in order to work properly, the decoder cannot act on its own, the work of extracting high level features from the characters surroundings is essential.

Let us drop FCNN out of the equation. By looking at the second experiment, we can observe a significant difference between the LSTM and its convolutional counterparts. Specifically, \hbox{Table \ref{tab:results-e2}} shows the accuracy of the convolutional encoders is about \SI{12}{\percent} higher than the one of the LSTM encoder. Nevertheless, this could have a simple explanation: it could be the case that the only thing that the LSTM needs is time. This leads us to the biggest differences between them: number of parameters, convergence time, and iteration time. Three points strengthen this argument: 
\begin{itemize}
    \item Table \ref{tab:parameters} shows the number of parameters of each model. The models only differ in the number parameters of the encoder (the rest of the model has \num{6368} million parameters), as expected. Thus, the LSTM encoder has ten times more parameters than the convolutional encoders, making it harder to train and more expensive to use.

    \item Figures \ref{fig:plot-e1} and \ref{fig:plot-e2} illustrate the convergence time differences. In particular, the LSTM encoder started to converge in E2 after \hbox{\SI{2}{\hour}\SI{45}{\min}} of training, whereas the convolutional encoders were close to their minimum at the mark of \hbox{\SI{1}{\hour}\SI{23}{\min}}.
    
    \item Regarding the iteration speed, \hbox{Tables \ref{tab:results-e1}} \hbox{and \ref{tab:results-e2}} show that, besides being more accurate, the convolutional encoders operate around two and four times faster than the LSTM encoder, respectively.
\end{itemize}

We present three arguments to explain this phenomenon: \hbox{(1) the} aforementioned difference in the number of parameters; \hbox{(2) the} existence of recursive connections in the LSTM, making it harder to optimize; and \hbox{(3) the} fact that convolutional networks run quite fast on GPUs. Anyway, this ensures that convolution-based encoders are viable, significantly faster, and statistically distinguishable from recurrent encoders (as shown in \hbox{Table \ref{tab:test-e2}}).

After solving one part of the second question, we just need to decide between the convolutional encoders. As show in \hbox{Table \ref{tab:results-e2}}, quantitatively both encoders are very similar, even though CFE obtains better results and is distinguishable from FE. Qualitatively, it looks quite brighter for the CFE encoder, for example:
\begin{itemize}
    \item The first example, \hbox{Figure \ref{fig:attn-mat-e1}}, shows that the three encoders behave in a similar fashion. However, CFE seems cleaner and more localized, since it knows better where to focus, to the point where it is easy to see three distinguished phases: day, month, and year.
    
    \item The second example, \hbox{Figure \ref{fig:attn-mat-e2}}, is clear. The LSTM encoder did not converge yet, so its prediction is way off the mark. Regarding the convolutional encoders, CFE gets the example right, its attention matrix seems quite clean, and it resembles a lot to an identity matrix; whereas the FE encoder struggles to maintain the focus (many non-diagonal elements have taken attention) and makes erratic leaps (which manifest in missing words in the prediction, see \hbox{Table \ref{tab:example-e2}}).
\end{itemize}

Thus, it can be concluded that, in this case, CFE is preferable to FE due to its qualitative benefits and, to a lesser extent, its quantitative results. Regarding the undetectable errors reported in \citet{DBLP:journals/corr/SproatJ16}, it can be firmly confirmed that they are not an issue in these models as they appear by chance due to solvable errors. Specifically, these errors are highly related with the attention mechanism as, like Table \ref{tab:errors} shows, the most common error is getting stuck in an infinite loop. These problems make the model lost its focus and jump around when confounding similar parts of the entry. Therefore, this could be greatly improved by using more sophisticated attention models that, for example, focus on local environments, take into account the index, or force the model to put more focus in the next character of the input.

Finally, we are going to discuss the results obtained on the third experiment. \hbox{Figure \ref{fig:plot-e3}} shows that, during training, the model quickly converged and there seems to be a gap between training and generalization error that the model has not been able to resolve. However, the results obtained on the test set are quite promising: it has obtained \SI{92.74}{\percent} accuracy and \SI{5.44}{\percent} CER, against the \SI{99.8}{\percent} accuracy and \SI{13.43}{\percent} CER obtained by the models of \citet{DBLP:journals/corr/SproatJ16} and \citet{DBLP:conf/aclnut/IkedaSM16}, respectively.

Leaving out the obvious differences (datasets, training time, and so on) that make comparing these models cumbersome, it is clear that these first results are promising and point out into a viable direction to solve the problem of text normalization in a data-driven fashion without the existence of undetectable errors being an unsolvable error, thus answering the first question made at the beginning of this section. 

\section{Conclusions}

In this paper, a new encoder-decoder architecture with attention mechanisms has been proposed for the problem of text normalization, using a character-level approach and introducing a new type of encoder. This encoder, called Causal Feature Extractor, is a brand new technique designed to work well in cooperation with the attention mechanisms. In the experiments, it has empirically proved to achieve good results, using the attention matrices more like someone would expect to use them by hand. Besides, it is able to work at least as good as the best of the compared encoders, and it brings all the benefits of using convolutional neural networks to the table. The last thing that distinguishes this encoder to the traditional recurrent encoders is its simplicity to be adapted to other input layouts (for example, matrices of pixels).

With respect to prior work, the initial results have shown to be quite close to the state-of-the-art, with plenty of room to future improvement. Despite getting a worse accuracy result (\SI{92.74}{\percent} vs. \SI{99.8}{\percent}), it does not critically suffer from undetectable errors (meaning that it can be viable for commercial use), nor it seems to concentrate its errors on any particular semiotic class since the errors are attention-based. 

Finally, some other interesting results can be extracted, like the introduction of a new variation of the attention mechanisms that uses a context matrix instead of a vector, or the empirical proof that empowers the role of encoders in the encoder-decoder architectures by showing that the system does not work if we just take features of single elements (without their neighborhood).

Future research lines could focus on some of the following:
\begin{itemize}
    \item The usage of the CFE encoder as a general-purpose encoder.
    \item A deeper training and hyperparameters selection to obtain better results.
    \item Exploring the errors made by the model, like the leap errors mentioned in \hbox{Section \ref{sec:discussion}}.
    \item Conditioning the model to external factors, for example, to distinguish between British and American English.
\end{itemize}

\section{Acknowledgments}

We would like to express our gratitude to Richard Sproat for his useful feedback on this article. Besides, Adri\'an acknowledges support from the Max Planck Institute for Intelligent Systems.

\starttwocolumn
\bibliography{references}

\begin{thebibliography}{11}
\expandafter\ifx\csname natexlab\endcsname\relax\def\natexlab#1{#1}\fi

\bibitem[{Bahdanau, Cho, and Bengio(2014)}]{DBLP:journals/corr/BahdanauCB14}
Bahdanau, Dzmitry, Kyunghyun Cho, and Yoshua Bengio. 2014.
\newblock Neural machine translation by jointly learning to align and
  translate.
\newblock \emph{CoRR}, abs/1409.0473.

\bibitem[{Chung, Cho, and Bengio(2016)}]{DBLP:conf/acl/ChungCB16}
Chung, Junyoung, Kyunghyun Cho, and Yoshua Bengio. 2016.
\newblock A character-level decoder without explicit segmentation for neural
  machine translation.
\newblock In \emph{Proceedings of the 54th Annual Meeting of the Association
  for Computational Linguistics, {ACL} 2016, August 7-12, 2016, Berlin,
  Germany, Volume 1: Long Papers}.

\bibitem[{Ikeda, Shindo, and Matsumoto(2016)}]{DBLP:conf/aclnut/IkedaSM16}
Ikeda, Taishi, Hiroyuki Shindo, and Yuji Matsumoto. 2016.
\newblock Japanese text normalization with encoder-decoder model.
\newblock In \emph{Proceedings of the 2nd Workshop on Noisy User-generated
  Text, NUT@COLING 2016, Osaka, Japan, December 11, 2016}, pages 129--137.

\bibitem[{Lee, Cho, and Hofmann(2017)}]{DBLP:journals/tacl/LeeCH17}
Lee, Jason, Kyunghyun Cho, and Thomas Hofmann. 2017.
\newblock Fully character-level neural machine translation without explicit
  segmentation.
\newblock \emph{{TACL}}, 5:365--378.

\bibitem[{van~den Oord et~al.(2016)van~den Oord, Dieleman, Zen, Simonyan,
  Vinyals, Graves, Kalchbrenner, Senior, and
  Kavukcuoglu}]{DBLP:conf/ssw/OordDZSVGKSK16}
van~den Oord, A{\"{a}}ron, Sander Dieleman, Heiga Zen, Karen Simonyan, Oriol
  Vinyals, Alex Graves, Nal Kalchbrenner, Andrew~W. Senior, and Koray
  Kavukcuoglu. 2016.
\newblock Wavenet: {A} generative model for raw audio.
\newblock In \emph{The 9th {ISCA} Speech Synthesis Workshop, Sunnyvale, CA,
  USA, 13-15 September 2016}, page 125.

\bibitem[{Riezler and III(2005)}]{DBLP:conf/acl/RiezlerM05}
Riezler, Stefan and John T.~Maxwell III. 2005.
\newblock On some pitfalls in automatic evaluation and significance testing for
  {MT}.
\newblock In \emph{Proceedings of the Workshop on Intrinsic and Extrinsic
  Evaluation Measures for Machine Translation and/or Summarization@ACL 2005,
  Ann Arbor, Michigan, USA, June 29, 2005}, pages 57--64.

\bibitem[{Sodimana et~al.(2018)Sodimana, Silva, Sproat, Theeraphol, Li, Gutkin,
  Sarin, and Pipatsrisawat}]{47344}
Sodimana, Keshan, Pasindu~De Silva, Richard Sproat, A~Theeraphol, Chen~Fang Li,
  Alexander Gutkin, Supheakmungkol Sarin, and Knot Pipatsrisawat. 2018.
\newblock Text normalization for bangla, khmer, nepali, javanese, sinhala, and
  sundanese tts systems.
\newblock In \emph{6th International Workshop on Spoken Language Technologies
  for Under-Resourced Languages (SLTU-2018)}, pages 147--151, 29-31 August
  2018, Gurugram, India.

\bibitem[{Sproat(1996)}]{DBLP:journals/nle/Sproat96}
Sproat, Richard. 1996.
\newblock Multilingual text analysis for text-to-speech synthesis.
\newblock \emph{Natural Language Engineering}, 2(4):369--380.

\bibitem[{Sproat and Jaitly(2016)}]{DBLP:journals/corr/SproatJ16}
Sproat, Richard and Navdeep Jaitly. 2016.
\newblock {RNN} approaches to text normalization: {A} challenge.
\newblock \emph{CoRR}, abs/1611.00068.

\bibitem[{Sutskever, Vinyals, and Le(2014)}]{DBLP:conf/nips/SutskeverVL14}
Sutskever, Ilya, Oriol Vinyals, and Quoc~V. Le. 2014.
\newblock Sequence to sequence learning with neural networks.
\newblock In \emph{Advances in Neural Information Processing Systems 27: Annual
  Conference on Neural Information Processing Systems 2014, December 8-13 2014,
  Montreal, Quebec, Canada}, pages 3104--3112.

\bibitem[{Xu et~al.(2015)Xu, Ba, Kiros, Cho, Courville, Salakhutdinov, Zemel,
  and Bengio}]{DBLP:conf/icml/XuBKCCSZB15}
Xu, Kelvin, Jimmy Ba, Ryan Kiros, Kyunghyun Cho, Aaron~C. Courville, Ruslan
  Salakhutdinov, Richard~S. Zemel, and Yoshua Bengio. 2015.
\newblock Show, attend and tell: Neural image caption generation with visual
  attention.
\newblock In \emph{Proceedings of the 32nd International Conference on Machine
  Learning, {ICML} 2015, Lille, France, 6-11 July 2015}, pages 2048--2057.

\end{thebibliography}

\end{document}